\documentclass[9pt,twocolumn,twoside]{opticajnl}
\journal{opticajournal} 

\setboolean{shortarticle}{true}
\usepackage{graphicx} 
\usepackage{subfigure} 
\usepackage{booktabs}
\usepackage{array}
\usepackage{amsmath}

\usepackage{lineno}
\title{Collimator-Assisted High-Precision Calibration Method for Event Cameras}
\author[1]{Zibin Liu\textsuperscript{$\dagger$,}}
\author[1]{Shunkun Liang\textsuperscript{$\dagger$,}}
\author[*,1]{Banglei Guan}
\author[1]{Dongcai Tan}
\author[1]{Yang Shang}
\author[1]{Qifeng Yu}
\affil[1]{College of Aerospace Science and Engineering, National University of Defense Technology, Changsha 410073, China}

\affil[$\dagger$]{These authors contributed equally to this work.}
\affil[*]{guanbanglei12@nudt.edu.cn}

\begin{abstract}
Event cameras are a new type of brain-inspired visual sensor with advantages such as high dynamic range and high temporal resolution. The geometric calibration of event cameras, which involves determining their intrinsic and extrinsic parameters, particularly in long-range measurement scenarios, remains a significant challenge. To address the dual requirements of long-distance and high-precision measurement, we propose an event camera calibration method utilizing a collimator with flickering star-based patterns. The proposed method first linearly solves camera parameters using the sphere motion model of the collimator, followed by nonlinear optimization to refine these parameters with high precision. Through comprehensive real-world experiments across varying conditions, we demonstrate that the proposed method consistently outperforms existing event camera calibration methods in terms of accuracy and reliability. 
\end{abstract}

\setboolean{displaycopyright}{false}

\begin{document}

\maketitle
As a novel type of brain-inspired visual sensor, event cameras represent a paradigm shift in visual sensing technology~\cite{Gehrig2024}. Unlike conventional cameras that capture frames at fixed intervals, these neuromorphic sensors asynchronously detect and report pixel-level brightness changes with microsecond temporal resolution. This unique operating principle endows event cameras with various advantages, including exceptional temporal resolution, high dynamic range, and resilience against motion blur~\cite{Zhou2024,10413505}. These characteristics are particularly suited for all-weather, long-distance observation tasks, such as space surveillance~\cite{9142352,10843758} and remote sensing~\cite{10928982,tao2023dudb}.

Prior to conducting observational tasks, camera calibration stands as an indispensable preliminary step~\cite{Kim2023,Huai2023}.
Existing event camera calibration methods can be classified into three principal categories. The first category of methods involves reconstructing images from event streams, utilizing algorithms such as E2VID~\cite{8946715} to convert these streams into intensity images, thereby aligning with conventional calibration workflows. The second category focuses on feature extraction from event streams, directly leveraging the inherent features within events for calibration, often employing geometric shapes such as circles~\cite{10555516} or lines~\cite{LIU2024114900}. Additionally, the third category relies on high-frequency flickering screens~\cite{prophesee2020} or LED light sources~\cite{dominguez2019bio}, thereby converting event streams into images. These methods are primarily optimized for indoor calibration. However, existing methods face significant challenges in long-range measurement applications due to signal deterioration over extended distances and interference from variable outdoor lighting conditions. Furthermore, the deployment of calibration targets becomes increasingly impractical in large field-of-view scenarios at considerable distances, complicating the calibration process. Without proper calibration, the unique advantages of event cameras, such as exceptional temporal resolution and high dynamic range, cannot be fully leveraged across various applications. The proposed method, also falling into this third category, is specifically designed to address these challenges inherent in long-range calibration.

\begin{figure}[t]
	\centering
	\includegraphics[width=0.45\textwidth]{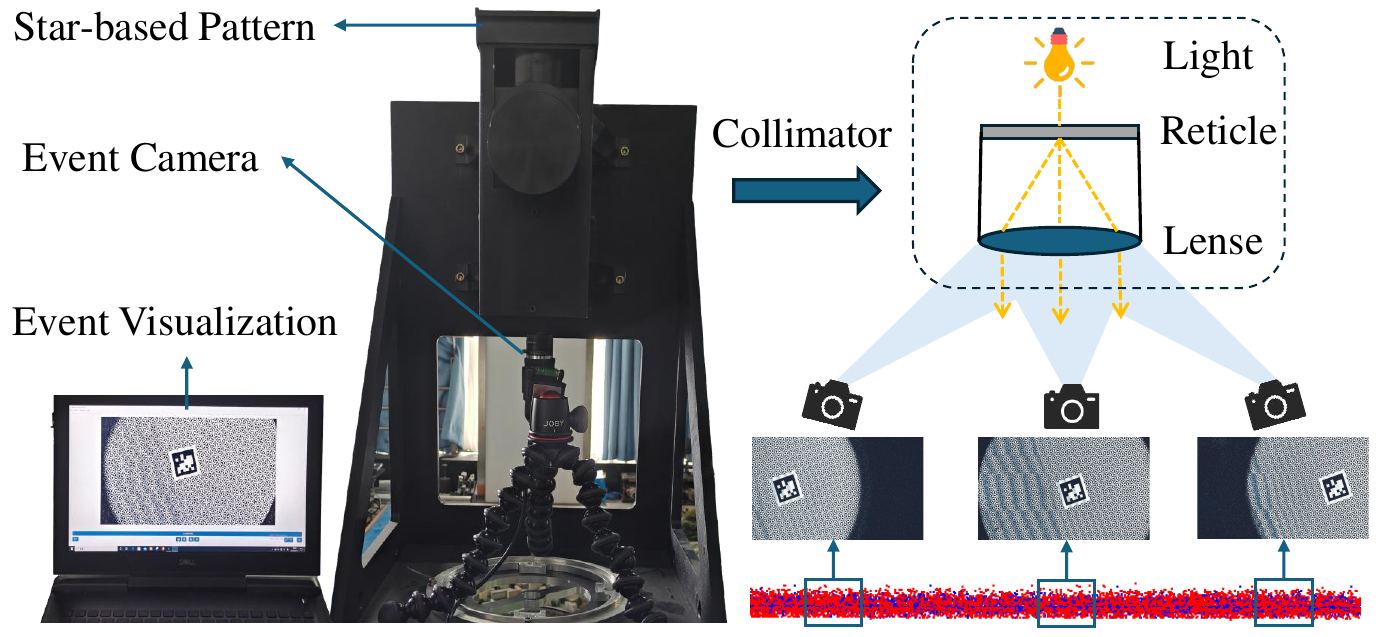}
	\caption{Collimator system and calibration principles.}
	\label{fig1}
\end{figure}

To address this issue, we employ a collimator system, which includes a collimator and flickering star-based patterns to achieve high-precision camera calibration. By collimating light into parallel beams, a collimator can simulate the imaging characteristics of targets at infinite distances. This capability offers several advantages when combined with event cameras. It enables controlled optical stimulation via a precisely regulated light source. As a result, consistent and repeatable event patterns are generated, unaffected by environmental conditions or interference from outdoor sunlight. Additionally, the collimator accurately replicates the characteristics of targets at optical infinity. This enables precise calibration for long-range applications without requiring extensive testing fields. Our method also simplifies the calibration process. It avoids the continuous motion of the camera or calibration targets for event generation, using flickering star patterns instead. This reduces calibration errors caused by feature extraction inaccuracies during movement, resulting in more reliable and accurate calibration results.

To the best of our knowledge, this is the first event camera calibration method that integrates a collimator system. The proposed method leverages both the imaging characteristics of the collimator and the flickering properties of star-based patterns. It can therefore simulate long-distance observation scenarios, enabling precise calibration of the geometric model for event cameras. Through rigorous experimental validation, we demonstrate the superior calibration accuracy and robustness of the proposed method compared to existing methods. The main contributions are: (\textit{i}) the proposal of an event camera calibration method that leverages the spherical motion model of the collimator for precise camera calibration, and (\textit{ii}) the design of a specialized collimator-based calibration system for event cameras that utilizes flickering star-based patterns, which generates clear events even when the camera and patterns remain stationary relative to each other.

Event cameras operate on a fundamentally different principle than traditional frame cameras, generating asynchronous events in response to local brightness changes in the scene. Despite this difference, the underlying imaging relationship of event cameras follows the standard pinhole camera model. When the brightness at a scene point $\mathbf{X}$ changes beyond a certain threshold, it triggers an event $\mathbf{e} = \{\mathbf{x}, t, \pm 1\}$ at the corresponding image location, where $\mathbf{x} = \left(x,y\right)^\top$ denotes the pixel coordinates of the event, $t$ is the timestamp of the event, and $\{\pm 1\}$ represents the event polarity, indicating a decrease or increase in brightness.

In the proposed calibration method, we position the event camera at one end of the collimator, with a star-based pattern placed at the other end, as shown in Fig~\ref{fig1}. By observing the star-based pattern from different angles, we can calibrate both the intrinsic and extrinsic parameters of the event camera. Without loss of generality, we set the Z-coordinate of the star-based pattern to zero in the world coordinate system. The transformation between the event camera image plane and the calibration target plane can be constrained by a $3 \times 3$ homography matrix $\mathbf{H}$~\cite{zhang_flexible_2000}:
\begin{equation}
\mathbf{H} = \lambda \mathbf{K} \left[ {\begin{array}{*{20}{c}} 
{{{\bf{r}}_1}}&{{{\bf{r}}_2}}&{\bf{t}}
\end{array}} \right],
\label{planar}
\end{equation}
where $\lambda$ is a scaling factor, and $\mathbf{r}_1, \mathbf{r}_2$ are the first and second columns of rotation matrix $\mathbf{R}$, and $\bf{t}$ is the translation vector. The matrix $\bf{K}$ includes the intrinsic parameters of the event camera, which include the focal lengths $\left( {{f_x},{f_y}} \right)$ and the principal point $\left( {{c_x},{c_y}} \right)$. Within the collimator, the relative motion between the calibration target and the camera satisfies the spherical motion model~\cite{liang2024camera}, meaning the original 6 Degrees-of-Freedom (DoF) general motion is constrained to a 3DoF pure rotation motion. The transformation from the event camera coordinate system $E$ to the calibration target coordinate system $P$ is defined by the rotation matrix $\mathbf{R}_{ep} \in SO(3)$ and the translation vector $\mathbf{t}_{ep}=[0,0,-r]\top$, where $r$ represents the radius of the spherical motion. The origin of the coordinate system $O$ is located in the upper left point of the calibration target. The relative pose between coordinate systems $P$ and $O$ is represented by the identity rotation matrix $\mathbf{R}_{po} = I$ and the translation vector $\mathbf{t}_{po}=[x,y,0]^\top$, indicating that the displacement occurs only in the $x$-$y$ plane of the calibration target. The transformation ${{\bf{T}}_{oe}}$ from the coordinate system $O$ to the event camera coordinate system $E$ can be expressed as:
\begin{equation}
\mathbf{T}_{oe} \!=\! \left( 
\begin{bmatrix}
\mathbf{I} & \mathbf{t}_{po} \\
\mathbf{0} & 1
\end{bmatrix} 
\begin{bmatrix}
\mathbf{R}_{ep} & \mathbf{t}_{ep} \\
\mathbf{0} & 1
\end{bmatrix} 
\right)^{-1} \!= \!
\begin{bmatrix}
\mathbf{R}_{ep}^{-1} & - \mathbf{R}_{ep}^{-1} (\mathbf{t}_{ep} + \mathbf{t}_{eo}) \\
\mathbf{0} & 1
\end{bmatrix},
\label{sphere}
\end{equation}%
where ${{\bf{t}}_{ep}} + {{\bf{t}}_{eo}} = [x,y, - r]^ \top$ is a fixed value. This constant vector represents a geometric invariant in the system that remains stable regardless of how the calibration target is rotated or moved, thus providing an important reference baseline for the calibration process. 

By integrating Eq. (\ref{planar}) with Eq. (\ref{sphere}), we obtain:
\begin{equation}
\frac{1}{\lambda}{{\mathbf{K}}^{-1}}\mathbf{H}=\left[ \begin{matrix}
   {{\mathbf{r}}_{1}} & {{\mathbf{r}}_{2}} & { - {\bf{R}}_{ep}^{ - 1}\left( {{{\bf{t}}_{ep}} + {{\bf{t}}_{eo}}} \right)}  \\
\end{matrix} \right].
\end{equation}

Exploiting the orthogonality property of the rotation matrix $\mathbf{R}$, it can be obtained:
\begin{equation}
\mathbf{H}^\top \mathbf{K}^{-\top} \mathbf{K}^{-1} \mathbf{H} = \lambda^2 \begin{bmatrix} 1 & 0 & -x \\ 0 & 1 & -y \\ -x & -y & {x^2} + {y^2} + {r^2} \end{bmatrix}.
\label{impt}
\end{equation}

By taking the inverse of Eq. (\ref{impt}) for $i$-th image, we have:
\begin{equation}
\mathbf{H}_i^{-1} \mathbf{K} \mathbf{K}^\top \mathbf{H}_i^{-\top} = \frac{1}{\lambda_i^2 r^2} \begin{bmatrix}
r^2 + x^2 & xy & x \\
xy & r^2 + y^2 & y \\
x & y & 1
\end{bmatrix}.
\label{last}
\end{equation}

This calibration framework first estimates seven parameters: four intrinsic parameters in $\mathbf{K}$ and three parameters defining the translation vector $(x, y, -r)^\top$. Each calibration image contributes five independent constraints through Eq. (\ref{last}), meaning that theoretically, a minimum of two images is sufficient to fully determine all unknown parameters in the system~\cite{liang2024camera}.

After obtaining initial estimates of the intrinsic and extrinsic parameters, further refinement is necessary to account for measurement errors and other factors that may affect accuracy. We implement a nonlinear optimization method that minimizes reprojection errors to achieve optimal camera parameters.

\begin{figure*}[t]
	\centering
	\includegraphics[width=0.9\textwidth]{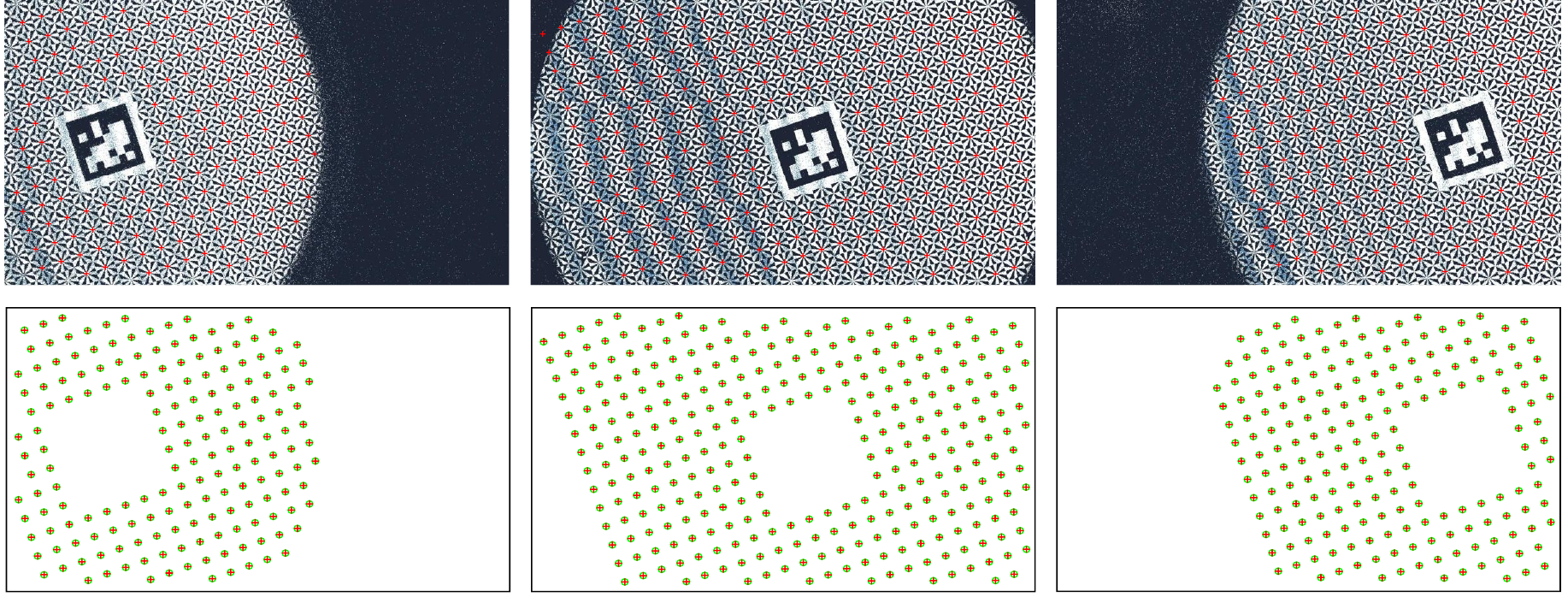}
	\caption{Event camera calibration process. The first row shows the marker point extraction results, with red crosses indicating the detected marker points. The second row displays the marker point reprojection results, with green circles representing the reprojected positions of the marker points.}
	\label{fig2}
\end{figure*}

Due to the inherent frequency-synchronized strobe light of the collimator, we obtain clear event-accumulated images. Given a set of $M$ event-accumulated images, each containing $N$ corresponding marker points, we formulate the optimization problem as:
\begin{equation}
\mathop {\min }\limits_\theta  \sum\limits_{i = 1}^M {\sum\limits_{j = 1}^N \rho  } \left( {{{\left\| {\boldsymbol{\pi} \left( {\theta ,{{\bf{X}}_{ij}}} \right) - {{\bf{x}}_{ij}}} \right\|}^2}} \right),
\end{equation}
where $\boldsymbol{\pi}$ is the projection function, and $\theta$ represents all the calibration parameters, including focal length, principal point, distortion coefficients, and extrinsic parameters. Without loss of generality, the initial values for distortion coefficients can be set to zero, and then solved through the optimization process~\cite{LIU2024114900}. $\left\{ {{{\bf{x}}_{ij}},{{\bf{X}}_{ij}}} \right\}$ represents the $j$-th corresponding 2D-3D point pair for the $i$-th calibration image. To mitigate the influence of potential outliers and noises on calibration quality, we incorporate the Huber loss function $\rho(\cdot)$, which behaves quadratically for small residuals and linearly for large residuals:
\begin{equation}
\rho(x) = 
\begin{cases}
x^2 & \text{if } |x| \leq \delta \\
2\delta|x| - \delta^2 & \text{otherwise}
\end{cases},
\end{equation}
where $\delta$ is a threshold parameter that controls the transition between the quadratic and linear regions.

We address this optimization problem using the Levenberg-Marquardt algorithm, which is a hybrid method that effectively merges the stability of gradient descent with the rapid convergence of Gauss-Newton. The refinement process concludes when either the maximum number of iterations is reached or when there is minimal improvement in the cost function, thereby ensuring optimal parameter estimation within the specified constraints. 

\begin{table*}[ht] 
	\centering
	\begin{tabular}{ccccc}
		\hline
		\textbf{Method}& \begin{tabular}[c]{@{}c@{}} \textbf{Prophesee Toolbox}~\cite{prophesee2020} \\ \end{tabular} & \textbf{Matlab Toolbox}~\cite{MatlabCalib} & \textbf{Zhang}~\cite{zhang_flexible_2000} & \textbf{Our} \\
		\hline
		\begin{tabular}[c]{@{}c@{}}Calibration target\end{tabular}&\begin{tabular}[c]{@{}c@{}}flickering screen \\ + Checkerboard pattern\end{tabular}  & \begin{tabular}[c]{@{}c@{}}flickering screen \\ + Checkerboard pattern\end{tabular} & \begin{tabular}[c]{@{}c@{}}Collimator \\ + Star-based pattern\end{tabular} &
		\begin{tabular}[c]{@{}c@{}}Collimator \\ + Star-based pattern\end{tabular} \\
		\hline
		\begin{tabular}[c]{@{}c@{}}Number of images\end{tabular}&20&20&3&3\\
		\hline
		$f_x$&3365.18 & 3337.61 & 3412.76 & 3345.06 \\
		\hline
		$f_y$& 3364.93& 3337.25 & 3412.75 & 3345.27 \\
		\hline
		$c_x$&633.21 & 624.07 & 647.06 & 642.10 \\
		\hline
		$c_y$ &362.05& 369.07 & 366.26 & 363.32 \\
		\hline
		$k_1$ &0.65&-0.07& 0.06 & 0.06 \\
		\hline
		$k_2$ &-1.39&5.43& -0.08 & -0.90\\
		\hline
		\begin{tabular}[c]{@{}c@{}}Reprojection error\end{tabular}&0.45&0.41& 0.10 & 0.08\\
		\hline
	\end{tabular}
	\caption{Camera calibration results. The intrinsic parameters $f_x$, $f_y$, $c_x$, $c_y$ and reprojection error are in pixels. The distortion coefficients $k_1$ and $k_2$ are dimensionless.}
	\label{tab1}
\end{table*}

\begin{figure}[t]
	\centering
	\includegraphics[width=0.4\textwidth]{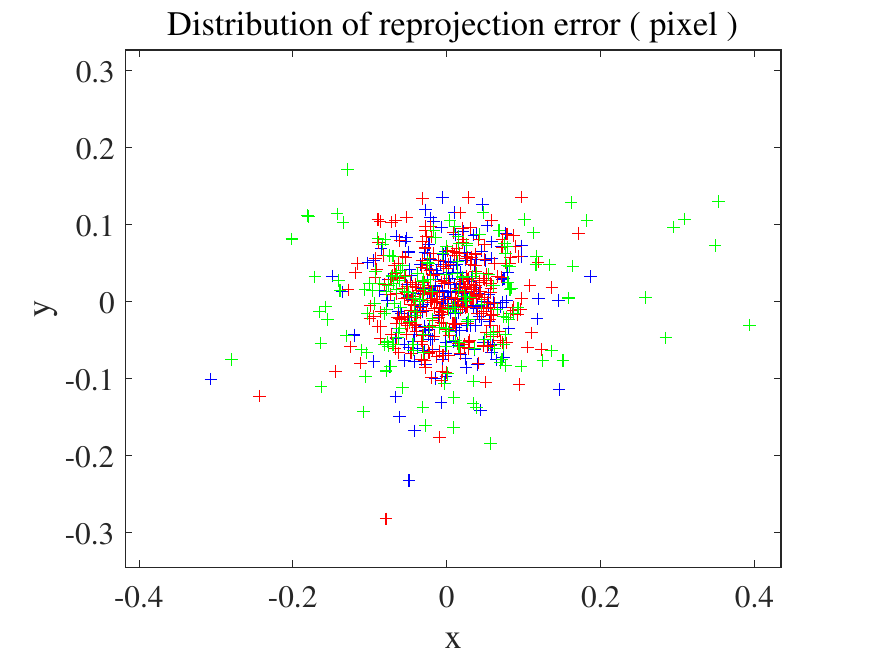}
	\caption{Distribution of marker point reprojection errors. Blue, green, and red represent the reprojection errors of marker points from the three event-accumulated images, corresponding to the left, middle, and right sub-figures of Fig. 2, respectively.}
	\label{fig3}
\end{figure}

In the experiment, we employ the Prophesee EVK4 event camera with a 16 $mm$ focal length and a resolution of 1280×720 $pixels$. A 60 Hz flickering planar light source, mounted at the top of the collimator, is used to generate events. The event streams are processed into event-accumulated images, generated by mapping event polarities to pixel values within 33 $ms$ intervals. This temporal resolution is selected to balance signal-to-noise ratio while preserving the dynamic characteristics of the events. We select three images for our calibration procedure to ensure statistical significance while maintaining computational efficiency. The calibration process consists of three primary stages. First, feature extraction~\cite{9156397} is performed on the event-accumulated image to identify the distinct marker points of the calibration pattern. Fig.~\ref{fig2} illustrates the results of the marker point extraction, where the red crosses indicate the extracted marker points. The results show that the majority of marker points have been accurately extracted. Following feature extraction, our algorithm performs initial parameter estimation to determine the intrinsic and extrinsic camera parameters. This step is essential as it provides the necessary initialization values for the subsequent optimization process. To refine these parameters, we implement an optimization framework using the Levenberg-Marquardt algorithm, which effectively minimizes reprojection errors across all event-accumulated images simultaneously, yielding optimal intrinsic and extrinsic camera parameters. As shown in Fig.~\ref{fig2}, the marker point reprojection results after optimization reveal high accuracy. The green circles indicate the reprojected marker points, which closely align with the extracted feature points. This visual representation provides intuitive verification of the calibration method's precision.

To validate the effectiveness of our calibration method, we compare our method with several established camera calibration methods. (\textit{i}) Prophesee Toolbox~\cite{prophesee2020} relies on a flickering screen with checkerboard pattern to generate event images. To ensure calibration accuracy, we selected 20 images for the calibration process. (\textit{ii}) The MATLAB Toolbox~\cite{MatlabCalib} also employs a flickering screen with checkerboard patterns and similarly inputted 20 calibration images. (\textit{iii}) Zhang’s method~\cite{zhang_flexible_2000} uses our collimator with star-based patterns as the calibration target. For a fair comparison, we provide the same three input images to both Zhang’s method and our algorithm. This ensures that any performance differences can be attributed to the algorithmic improvements rather than variations in input data or calibration targets.

The quantitative performance of our calibration method is summarized in Table~\ref{tab1}, showing significant improvements in calibration accuracy compared to conventional methods. Our method achieves a mean reprojection error of 0.08 pixels, with the error distribution shown in Fig.~\ref{fig3}, which represents the state-of-the-art performance compared to standard calibration techniques applied to event cameras. A key advantage of our method is its efficiency. Our collimator-based method achieves exceptional calibration precision using only three images, while conventional methods typically require more calibration images to achieve acceptable results. For instance, the Matlab Toolbox with checkerboard patterns demands 20 calibration images yet still produces substantially larger reprojection errors compared to our method. This dramatic reduction in required calibration data, offers compelling practical benefits by significantly streamlining the calibration workflow. More importantly, these results experimentally validate the theoretical advantages of employing a collimator for long-distance calibration of event cameras, a scenario where traditional approaches typically struggle to maintain accuracy. The ability to achieve superior calibration with fewer images also demonstrates the inherent stability and robustness of our method, making it particularly valuable for applications where calibration opportunities may be limited or time-constrained. When comparing our method with Zhang's method, both implemented using the same star-pattern collimator and identical three input images, our method demonstrates a slight precision advantage in terms of reprojection error. Notably, the intrinsic parameters obtained from our method are closer to those derived from multi-image calibration methods. This finding is significant as it further demonstrates that the advantages of our method extend beyond just the use of a collimator, they also stem from the robustness of our proposed algorithmic framework. This superior performance under identical input conditions reinforces the applicability and reliability of our method in the high-precision calibration of event cameras.

In conclusion, this work proposes a novel event camera calibration method that leverages a designed collimator system. Based on the spherical motion model of the collimator system, the proposed method first linearly solves for the event camera's parameters. It then further optimizes the camera parameters through nonlinear optimization. The experimental results demonstrate that the proposed method achieves significantly higher calibration accuracy compared to traditional calibration methods, while requiring fewer images. Additionally, the proposed method is not limited to event cameras, but can also be extended to frame-based cameras. By providing more accurate and robust camera calibration, this work could unlock more applications for event cameras in astronomy, remote sensing, and other fields that require precise visual perception at extended ranges or under demanding environmental conditions. 

\begin{backmatter}
\bmsection{Funding} National Natural Science Foundation of China under Grant 12372189, and the Hunan Provincial Natural Science Foundation for Excellent Young Scholars under Grant 2023JJ20045.

\bmsection{Disclosures} The authors declare no conflicts of interest.

\bmsection{Data availability} Data underlying the results presented in this paper are not publicly available at this time but may be obtained from the authors upon reasonable request.
\end{backmatter}

\bibliography{main}

\end{document}